\documentclass{iopjournal}

\usepackage[numbers,sort&compress]{natbib}
\usepackage{csquotes}
\usepackage{booktabs}
\usepackage{graphicx}
\usepackage{amsmath}
\usepackage{amssymb}
\usepackage{multirow}
\usepackage{siunitx}
\newcommand{\pmac}{\ensuremath{P_{\mathrm{macro}}}}   

\begin{document}

\title{Accountable and uncertainty-aware evaluation of sensor-based AI under distribution shift: devices, subjects, and nearly three years underground}

\author{Benny Platte$^{1,*}$, Rico Thomanek$^{1}$, Christian Roschke$^{1}$ and Marc Ritter$^{1}$}

\affil{$^1$Faculty of Applied Computer Sciences and Biosciences, Mittweida University of Applied Sciences, Mittweida, Germany}

\affil{ORCID: Benny Platte 0000-0001-7754-5170, Rico Thomanek 0009-0000-6906-6209, Christian Roschke 0009-0007-2875-5199, Marc Ritter 0009-0004-0204-8275}

\affil{$^*$Author to whom any correspondence should be addressed.}

\email{benny.platte@hs-mittweida.de}

\keywords{indoor positioning, geomagnetic fingerprinting, distribution shift, measurement uncertainty, evaluation protocol, accountability, LSTM}

\begin{abstract}
Sensor-based AI systems are rarely operated under the conditions under which they were trained: devices, personnel and recording epochs change, and each change degrades performance in ways a random train--test split cannot reveal. We propose a staged, accountable evaluation protocol that treats the evaluation of a deployed model as a measurement with declared reference levels and a quantified uncertainty. Four cumulative generalisation stages hold out devices, subjects and time. Each stage is judged on quantiles of repeated trainings against chance references with the correct class count, an out-of-present-scope rate exposes silent misdirection towards classes that are no longer present in deployment relative to training, and an explicit decision rule ties roll-out decisions not to means but to 5\% quantiles. We demonstrate the protocol on infrastructure-free geomagnetic localisation with smartphone-based recurrent classifiers in two real underground mines, including a replication of the scheme's training stages at the second site. Unchanged models are re-evaluated on data recorded 34 months after the training campaign, on a device generation unknown at training time and with a held-out surveyor. The $5\%$ quantile of their present-conditioned precision there is $0.39$ over 299 repeated trainings, $16.5$ times the chance level; across the composition of the 42 reachable location classes the figure varies by $\pm0.08$, several times the spread between repeated runs. Repeated trainings of a single configuration show why means mislead: a bimodal configuration passes a mean-based test decisively while its $5\%$ quantile lies more than an order of magnitude below chance.

\end{abstract}

\section{Introduction}
\label{sec:intro}

A learned model learned on sensor data changes its measurement conditions when it leaves the laboratory. The sensor fleet, in the present case the smartphone fleet, ages and is replaced, personnel changes, and the deployment outlives the recording campaign that produced the training data by years. Each of these changes is a distribution shift between training and operation \cite{QuinoneroCandela_ea_2009_DatasetShiftInMachine, Koh_ea_2021_WILDSBenchmarkOfIntheWild}, and none of them is visible in the evaluation practice that dominates the fingerprint-localisation literature, where a single campaign is split into training and test portions and the resulting accuracy is reported as the performance of the system (section~\ref{sec:related:geomag}).

This paper treats the problem as one of measurement science. If the performance of a deployed model is a measured quantity, then its report owes the reader the same elements that the \enquote{Guide to the Expression of Uncertainty in Measurement} demands of any measurement \cite{JointCommitteeforGuidesinMetrology_2008_EvaluationOfMeasurementData, JointCommitteeforGuidesinMetrology_2023_GuideToTheExpression, 2026_GuideToTheExpressionInMeasurement}: a defined measurand, declared reference levels, and a quantitative statement of uncertainty. The protocol proposed here supplies these elements for sensor-based classification systems. Its stages hold out, cumulatively, the devices, the subjects and the recording epoch; its reference levels are chance lines with the correct class count, a self-referenced plausibility bound for predictions on classes that are no longer present in deployment relative to training, and the random split, which is demoted to a declared upper bound; its uncertainty statement rests on repeat trainings of identical configurations, summarised by quantiles because the relevant risk of a deployment is the weak single draw rather than the mean.

The protocol is demonstrated end to end on a hard instance: infrastructure-free localisation that uses the geomagnetic field as its only external signal source, recorded with devices that would plausibly be at hand in a disorientation situation in mines or caves, namely ordinary smartphones. Satellite navigation and radio infrastructure of any kind are unavailable. The only signal is the local (micro-)distortion of the geomagnetic field, that is micro-anomalies, which is of a similar order of magnitude to the sensor error (section~\ref{sec:setting}). The demonstration spans two structurally different underground mines, with 7003 training runs used to produce the models, and includes what we believe to be the first re-evaluation of unchanged geomagnetic signature-localisation models on data recorded years after training: 34 months, on a device generation released after the training campaign, by a surveyor the models had never seen in training (section~\ref{sec:stage4-longtime}).

\paragraph{Contributions.}
\begin{enumerate}
\item A staged evaluation protocol with declared reference levels and an explicit, quantile-based decision rule (section~\ref{sec:protocol}), applied retrospectively to all generalisation stages of both mines (section~\ref{sec:verdictsub}).
\item A quantification of the (over-)optimism of the random split on matched configurations, $+0.14$ in the cross-device scenario (device change) and $+0.41$ in the cross-subject scenario (subject change), together with its leakage mechanism through window overlap and the resulting reporting rule (sections~\ref{sec:optimism_gap} and~\ref{sec:leakage}).
\item A cross-time evaluation over 34 months with a three-part decomposition of the degradation into class loss, annotation-boundary ambiguity and a bounded residual of genuine signature change, guarded by an out-of-present-scope (OOPS) rate (section~\ref{sec:stage4-longtime}).
\item A three-step one-shot risk assessment over repeated trainings that evidences deployment-relevant lower bounds and demonstrates, on a bimodal configuration, that mean-based reporting misclassifies deployment risk (section~\ref{sec:uncertainty}).
\item Representation findings with protocol relevance: augmenting the model input with additional attitude information hurts under shift, and the ranking of representations from the random split does not predict their ranking under the escalated generalisation demands of the deployment stages (section~\ref{sec:representations}).
\end{enumerate}

This paper works on model clouds from the campaigns reported here: the point clouds that the fully recorded training runs of one campaign under one evaluation protocol form in configuration--performance space, as defined in \cite{platte2026companion}. The term is geometric and unrelated to cloud computing. Explanation is used here in the accountability sense of assigning observed performance loss to identifiable causes and producing the evidence a deploying actor owes to a reviewing forum \cite[p.~450]{Bovens_2007_AnalysingAndAssessingAccountability}\,\cite[p.~10]{Wieringa_2020_WhatToAccountFor}.

\section{Related work}\label{sec:related}

\subsection{Geomagnetic indoor localisation with deep learning}\label{sec:related:geomag}
In mines and cave systems, external localisation systems such as GPS or radio positioning are unavailable. Geological structures, for example in the form of ore veins, distort the local geomagnetic field and produce position-dependent signatures. A smartphone magnetometer records these signatures during a walk, without installed infrastructure. Xie et al.\ demonstrated magnetic fingerprinting on smartphones for recognising positions, with an augmented particle filter \cite{Xie_ea2014MaLocPracticalMagneticFingerprintingApproach}. Deep learning entered the field with DeepML, an LSTM \cite{Hochreiter_Schmidhuber_1997_LongShortTermMemory} trained on magnetic and light intensity data \cite{Wang_ea_2018_LSTMForIndoorPosMagnetic}, and AMID, a deep neural network that classifies magnetic landmarks \cite{Lee_ea2018AMIDAccurateMagneticIndoorLocalization}. Subsequent work classifies indoor landmarks from geomagnetic sequences with recurrent networks \cite{Bhattarai_ea_2019_GeomagneticFieldBasedIndoor}, learns magnetic patterns with convolutional networks \cite{Ashraf_ea2020MINLOCMagneticFieldPatternsBasedIndoor}, and combines residual, transformer, and LSTM blocks for fingerprint positioning in complex indoor environments \cite{Guo_2025_RobustMagneticFingerprintPositioning}. In measurement science, deep magneto-inertial fusion has recently been applied to road localisation \cite{Luo_ea_2026_MMIFDLRLMultimodalMagnetoinertialFusionbased}. However, underground studies are scarce. Makkonen et al.\ established the feasibility of smartphone-based geomagnetic positioning in a tunnel \cite{Makkonen_ea_2014_TheApplicabilityOfGeomagnetic}, Wang et al.\ tested geomagnetic matching algorithms for underground positioning in mine tunnels, recorded with a portable fluxgate magnetometer \cite{Wang_ea_2019_PerformanceTestOfMPMD}, and Seguel et al.\ survey positioning technologies for underground mines \cite{Seguel_ea_2022_UndergroundMinePositioningReview}.

These studies share one evaluation pattern: data from a single measurement campaign are split into training and test portions, and the reported accuracy therefore refers to the devices, surveyors, site, and period seen during training. Surveys of magnetic indoor positioning identify device heterogeneity as a persistent obstacle, since magnetometers in different smartphones differ in bias, sensitivity and noise characteristics and therefore record systematically different field values at the same location \cite{He_Shin2017GeomagnetismForSmartphoneBasedIndoorLocalization, Ashraf_ea2020ComprehensiveAnalysisOfMagneticField}. Where the fingerprinting literature addresses heterogeneity, it does so by calibration within the campaign \cite{Xie_ea2014MaLocPracticalMagneticFingerprintingApproach}. Performance on a device generation, a person, or a measurement epoch that is entirely absent from training remains unreported in this literature. The protocol proposed in this paper makes these under-reported factors the measurand.

\subsection{Long-term fingerprinting}\label{sec:related:longterm}
The temporal stability of fingerprints has been studied mainly for radio-based localisation, in particular WiFi. Mendoza-Silva et al.\ published a dedicated long-term dataset whose collection spanned 15 months, with monthly training and test sets designed for research on signal ageing and robust positioning \cite{MendozaSilva_ea_2018_LongTermWiFiFingerprintingDataset}. This line quantifies radio-map ageing and keeps the option of periodic retraining open. For the geomagnetic modality we are not aware of a comparable longitudinal resource. In particular, we found no published re-evaluation of unchanged geomagnetic models on data recorded years after training, on a device generation released after training, and with a surveyor absent from the training data. The re-evaluation reported here covers 34~months under exactly these conditions.

\subsection{Evaluation under distribution shift and measurement uncertainty}\label{sec:related:shift}
Distribution shift between training and operating conditions is an established failure mode of learned models \cite{QuinoneroCandela_ea_2009_DatasetShiftInMachine}, and taxonomies distinguish covariate shift, prior probability shift, and concept shift \cite{MorenoTorres_ea_2012_UnifyingViewOnDataset}. The WILDS benchmark documents across ten datasets that in-distribution test accuracy systematically overstates performance under naturally occurring shifts across sites, devices, and time \cite{Koh_ea_2021_WILDSBenchmarkOfIntheWild}. Sensor-based localisation is exposed to the same shift axes through device fleets, changing personnel, and long operating periods, yet the fingerprinting literature reviewed above evaluates in distribution. For localisation systems, ISO/IEC~18305 already provides a test and evaluation framework that prescribes test environments and scenarios \cite{InternationalOrganizationforStandardization_InternationalElectrotechnicalCommission_2016_InformationTechnologyRealTime}; it does not address how a learned model is to be assessed across device fleets, personnel and years of operation. Two works from machine learning supply the missing pieces. Bouthillier et al.\ show that the spread of repeated training runs regularly overturns the ranking of methods, and derive from this the requirement of repeat ensembles rather than single runs \cite{Bouthillier_ea_2021_AccountingForVarianceIn}. Kapoor and Narayanan show across eighteen fields that data leakage is the most common cause of irreproducible results, and name the splitting of dependent observations explicitly as a leakage type \cite{Kapoor_Narayanan_2023_LeakageAndTheReproducibility}. That is precisely the case for overlapping windows of the same time series (section~\ref{sec:leakage}). For the reporting side, measurement science provides an established reference: the \enquote{Guide to the Expression of Uncertainty in Measurement} requires every measurement result to carry a quantitative statement of its uncertainty \cite{JointCommitteeforGuidesinMetrology_2008_EvaluationOfMeasurementData, JointCommitteeforGuidesinMetrology_2023_GuideToTheExpression, 2026_GuideToTheExpressionInMeasurement}. This paper treats the evaluation of a deployed model as a measurement in this sense. Each protocol stage reports its result against declared reference levels, and the one-shot risk assessment states quantiles of the performance distribution rather than a single mean value.

\subsection{Accountability and the role of explanation}\label{sec:related:accountability}
Bovens defines accountability as a relation between an actor and a forum: the actor is obliged to explain and justify conduct, the forum can pose questions and pass judgment, and the actor may face consequences \cite[p.~450]{Bovens_2007_AnalysingAndAssessingAccountability}. Wieringa transfers this framework to algorithmic systems and concludes from a systematic review that accountability must be established separately for each phase of a system life cycle \cite[p.~10]{Wieringa_2020_WhatToAccountFor}. Information systems research has mapped explainability objectives to stakeholder groups \cite{Meske_ea_2022_ExplainableArtificialIntelligenceObjectives, Bauer_ea_2021_ExplAInItToMe}. For time series models, Theissler et al.\ review explainability methods, which operate on model internals and input attributions \cite{Theissler_ea_2022_ExplainableAIForTime}. The present paper uses explanation in the accountability sense. The staged protocol produces the evidence a deploying actor owes to a forum, and the degradation decomposition assigns observed performance loss to identifiable causes, namely signature loss and annotation artefacts. Attribution-based explanation of the model configuration for the same case study is the subject of a foundational paper \cite{platte2026companion}. Recent fingerprinting work names reliability under real-world conditions as the open problem while evaluating on static benchmark datasets \cite{Ayinla_ea_2026_RELocAnEnhanced3D}. The protocol presented here supplies an evaluation instrument for this problem.

\section{Experimental setting}\label{sec:setting}

\subsection{Measurement campaigns}
\label{sec:methods:campaigns}

The evaluation protocol was instantiated in three measurement campaigns in two underground sites whose magnetic environments differ in origin and structure. The first site is a historic adit of the old silver mining district, the Markus-R\"ohling-Stolln in Frohnau (Annaberg-Buchholz, Ore Mountains, Germany). Narrow, hand-cut galleries follow the ore veins, and cobalt and nickel bearing minerals together with sulfide-bearing bands imprint location-specific distortions on the geomagnetic field. The walked total route of about \qty{1.2}{\kilo\metre} is derived from the nominal marker spacing; no surveyed total length is documented. The site was instrumented with 47 linear markers at a nominal spacing of \qty{25}{\metre}. The second site is gallery~35 of the former uranium mine Schlema-Alberoda (SAG/SDAG Wismut, Bad Schlema, Germany), in operation until 1991. The measurement route covers about \qty{1.5}{\kilo\metre} including the side galleries 806 and 809, with a nearly straight main gallery of about \qty{800}{\metre}. Large gallery cross-sections dominate the magnetic environment there. The second site therefore serves as a structural replication of the complete evaluation scheme under a different distortion regime.

The first campaign took place in the historic mine. Three smartphones (table~\ref{tab:devices}) recorded the magnetic field vector, device attitude and gravity at \qty{50}{\hertz}. They sat in a bracket that reproduces the natural one-handed carrying posture and is carried in the hand like a single phone. It is rigid only in that it fixes the orientation of the devices relative to each other, so that all devices record in the same attitude at the same time. Two operators walked the accessible section in eleven recording walks of \qty{10}{\minute} to \qty{1}{\hour} each. The section was equipped with 49 numbered location markers (numbers 1--47, 50 and 51). The target class is the combination of marker and movement direction: 46 markers were traversed in both directions, yielding 92 direction-resolved classes, and three markers (47, 50 and 51) were annotated as dwell locations without a movement direction, resulting in 95 classes in total. After merging and cleaning, the dataset comprises 113 gap-free sequences with \num{816161} samples. A later re-cleaned state of the same dataset, on which the training logs evaluated here rest, contains 110 sequences with \num{804543} samples. On this dataset the four evaluation stages of the protocol were populated from three model clouds of 1306 random-split models (the declared upper bound), 416 cross-device and 504 cross-subject models, complemented by 1072 repeated runs of identical configurations for the uncertainty analysis.

\begin{table}
\caption{Recording devices. \emph{Primary} denotes the device on which the operator annotated the location markers live, \emph{secondary} the co-carried devices without annotation. In campaign~2 operator~1 carried the three older devices and operator~2 the iPhone~16~Pro and the iPhone~13. Sensor figures are taken from the data sheets of the built-in components as far as the manufacturers state them; \enquote{n.\,s.} means not stated by the manufacturer.}
\label{tab:devices}
\centering\small
\begin{tabular}{llll}
\toprule
Device & Campaign & Role & Resolution / noise \\
\midrule
iPhone~14~Pro & 1, 2 & primary   & \qty{0.3}{\micro\tesla}/LSB; \qty{0.3}{\micro\tesla} \\
iPhone~X      & 1, 2 & secondary & n.\,s. \\
iPhone~7      & 1, 2 & secondary & \qty{0.15}{\micro\tesla}/LSB \\
iPhone~16~Pro & 2, 3 & primary   & \qty{0.16}{\micro\tesla} (SD); \qtylist{0.19;0.45}{\micro\tesla} rms$^{\ast}$ \\
iPhone~13     & 2    & secondary & \qty{0.22}{\micro\tesla} (SD) \\
\bottomrule
\end{tabular}

\vspace{0.5ex}
{\footnotesize $^{\ast}$ Noise figures of the iPhone~16~Pro stated separately for the $x$, $y$ and the $z$ axis.}
\end{table}

The second campaign took place in the uranium mine and repeated the scheme with an enlarged device pool. Five smartphones spanning five hardware generations (table~\ref{tab:devices}) were distributed over two operators, one carrying three devices and one carrying two. This assignment remained fixed throughout the campaign. Device and person are therefore not fully separable in the uranium mine: a device change across the carrier boundary is at the same time a change of person. Which device and which person the splits held out is not recorded in the archived training logs. Twenty-one markers were placed at intervals of \qty{50}{\metre} to \qty{100}{\metre}, oriented at the official gallery chainage marks and at salient features such as crossings and support changes; the final marker was discarded automatically because the adjacent section was too short to accommodate the longest segments investigated during segment formation. The remaining 20 markers in two directions define 40 classes. The merged database contains \num{2507658} samples, corresponding to \num{50153} seconds of recording across all devices. Sequence sections shorter than 700 samples were removed (99 of 1138 sections). This campaign yielded three model clouds of 1185 random-split, 1152 cross-device and 1152 cross-subject models, together with 216 repeated runs of identical configurations, of which the three configurations with at least nine repeats enter the uncertainty analysis.

The third campaign is a re-measurement in the historic mine, 34~months after the training campaign, and provides the data for the cross-time stage. All models remained unchanged; the campaign is a pure inference setting. Access was restricted to the rear loop of the mine beginning at marker~24, so that 21 of the original markers (24, 25--28 and 31--46) and hence 42 of the 95 trained classes were reachable. The physical markers had been removed after the first campaign and were reconstructed from the planning documents and from salient structures such as machines, junctions and narrowings. Since many positions of the first campaign had been set by step counting, exact reproduction was impossible, and boundary shifts between adjacent classes are an expected annotation artefact of the re-measurement. A further harmonisation step concerned the movement direction: the loop was walked as one continuous sequence, so markers 28 to 24 were traversed opposite to the coding of the first campaign, and the direction labels of markers 24--28 were inverted and shifted by one position to restore the label semantics of the training data. The recording device was an iPhone~16~Pro, which was not yet on the market at the time of the training campaign, and the operator was the person whose recordings from the training campaign had served exclusively as validation data in the cross-subject stage. The cross-time stage thus combines an unseen device, an unseen person and a temporal gap of nearly three years in a single setting. Two complete walks of the accessible loop yielded \num{60709} samples at \qty{50}{\hertz}.

\begin{figure}[t]
\centering
\includegraphics[width=\linewidth]{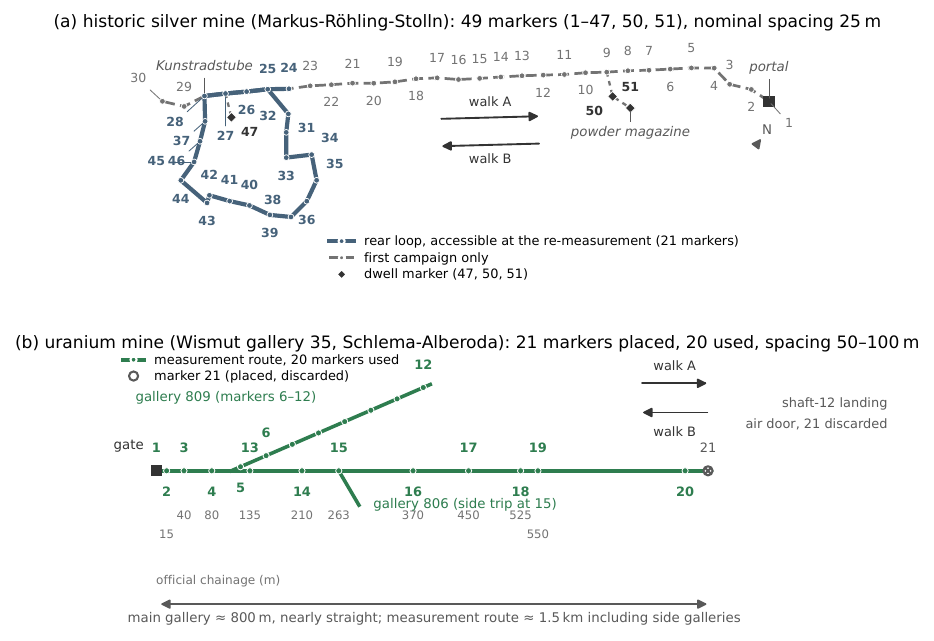}
\caption{Topology of the two measurement sites. (a)~Historic silver mine: 49 markers (1--47, 50, 51) at a nominal spacing of \qty{25}{\metre}; marker positions are taken from the official operating plan of the show mine, and the map frame is rotated by \ang{37} so that the main gallery runs horizontally; the loop accessible in the re-measurement (markers 24, 25--28, 31--46) is drawn in solid colour, sections recorded only in the first campaign in dashed grey; dwell markers are shown as diamonds; arrows give the two walking directions. (b)~Uranium mine, gallery 35, schematic: straight main gallery of about \qty{800}{\metre} with side galleries 806 and 809 (about \qty{1.5}{\kilo\metre} in total); 21 markers at \qtyrange{50}{100}{\metre} spacing, of which marker 21 was discarded.}
\label{fig:sites}
\end{figure}

Figure~\ref{fig:sites} sketches the topology of both routes. A class corresponds to a direction-resolved gallery segment of nominally \qty{25}{\metre} in the historic mine and \qtyrange{50}{100}{\metre} in the uranium mine. A window of 700 samples at \qty{50}{\hertz} spans \qty{14}{\second}, at walking speed about \qtyrange{15}{20}{\metre} and thus more than half a segment. A class-exact hit therefore localises the user to one such segment including walking direction, and the one-neighbour tolerance of the cross-time stage corresponds to one marker spacing.

The measurement task operates close to the sensor noise floor. The geomagnetic total field in central Europe amounts to \qtyrange{48}{50}{\micro\tesla}, whereas the local anomalies that carry the position information lie between \qty{-0.72}{\micro\tesla} and \qty{1.21}{\micro\tesla} \cite{Gabriel_ea_1980_AnomaliesOfTheEarths}. Datasheet-derived error bounds of the smartphone magnetometers range from \qty{\pm 0.1}{\micro\tesla} to \qty{\pm 0.45}{\micro\tesla} and are therefore of the same order of magnitude as the signal of interest. The recording device of the re-measurement sits at the upper end of that range with \qty{0.45}{\micro\tesla} rms on the $z$ axis, so a contribution of the device characteristic to the cross-time degradation cannot be excluded; it is contained in the residual of the decomposition in section~\ref{sec:stage4-longtime}. This proximity of signal and error scales motivates the sequence-based classification approach and the uncertainty-aware reporting used throughout the protocol.

\subsection{Ground truth and synchronisation}
\label{sec:methods:groundtruth}

Ground truth follows a leader--follower paradigm: a reference walk produces annotated training data, and a later user applies the trained models without individual calibration. Annotation was performed live on one designated primary device per operator. In the first campaign the walker selected the next marker approximately midway between markers; in the second campaign the switch was made at the moment a marker became visible. Each recorded sample therefore carries the currently active marker. In the uranium mine the two operators walked separately, partly at large distance, and decided independently when to switch markers, which keeps the annotations of different persons statistically independent. Marker positions in the historic mine were planned from the mine survey plan at a spacing of \qty{25}{\metre} and could largely be placed at the planned locations. Individual markers were moved by up to about \qty{10}{\metre} to salient structures such as branches, constrictions and machinery; between markers 13 and 15 this produced a section of about \qty{50}{\metre}. That is documented and acceptable because the protocol identifies places and metric coordinates play no role for the task. The \qty{25}{\metre} quoted below are therefore to be read as a nominal figure.

Synchronisation across devices proceeds in two stages. Device clocks were aligned before each campaign, and every continuous recording block carries a globally unique session identifier, which allows offline merging without any inter-device communication underground. Without further correction, timestamp deviations between devices ranged from \qty{0}{\second} to \qty{0.21}{\second}, with the majority below \qty{50}{\milli\second}. Sample-accurate alignment was then obtained from signal edges generated by the attitude-controlled recording logic: tilting the device between the steep carrying position and the measuring position pauses and resumes the recording and imprints a characteristic edge on the time series at every start. Annotations of the primary device were propagated to the secondary devices by nearest-timestamp matching with a tolerance of \qty{2}{\second}. Sequences were defined as runs of consecutive samples with timestamp gaps of at most \qty{50}{\milli\second}, permitting one dropped sample at the nominal spacing of \qty{20}{\milli\second}; in the first campaign, sequences shorter than 200~samples (\qty{4}{\second}) were removed, reducing the identified sequences to the 113 retained ones. The same pipeline was applied in the later campaigns.

\section{A staged evaluation protocol under distribution shift}
\label{sec:protocol}

\subsection{Stages and measurand}
\label{sec:protocol:stages}

The protocol treats the evaluation of a deployed model as a measurement. The measurand is the macro-averaged precision \pmac{} of the location classification, chosen because the class distribution follows the gallery topology and is imbalanced; macro averaging weights every location equally and the chance level under uniform random prediction is $1/K$ for $K$ classes irrespective of the imbalance. Four cumulative stages instantiate the shift axes under which a deployed system actually operates:

\begin{enumerate}
\item \textbf{Random split.} Training and validation windows are drawn from the same recordings. This stage enters the protocol solely as a declared upper bound (section~\ref{sec:optimism_gap}) and is always reported together with its window overlap (section~\ref{sec:leakage}).
\item \textbf{Cross-device.} Validation data come from a smartphone that contributed no training data. This is the fleet question: does the model survive a device change.
\item \textbf{Cross-subject.} Validation data come from a person absent from the training data, recorded on a device that is also held out. This is the personnel question.
\item \textbf{Cross-time.} The unchanged models are re-evaluated on recordings made years later, on a device generation that did not exist at training time, by the held-out person. This is the operating-period question, and it accumulates the previous axes.
\end{enumerate}

Formally, for a frozen model $m$ evaluated under conditions $B$ (device fleet, surveyor, recording epoch, site) the measurand of stage $s$ is
\begin{equation}
\pmac(m; B_s) \;=\; \frac{1}{|\mathcal{K}_s|} \sum_{k \in \mathcal{K}_s}
\frac{\mathrm{TP}_k}{\mathrm{TP}_k + \mathrm{FP}_k},
\label{eq:pmac}
\end{equation}
where $\mathcal{K}_s$ is the class inventory of the stage and a class that the model never predicts ($\mathrm{TP}_k+\mathrm{FP}_k=0$) contributes zero. Each stage therefore defines its own measurand: the same quantity under different, declared conditions $B_s$. In the vocabulary of the International Vocabulary of Metrology \cite{JointCommitteeforGuidesinMetrology_2012_InternationalVocabularyOfMetrology} the repeat ensembles of section~\ref{sec:uncertainty}, that is, the sets of repeated trainings of one fixed configuration on identical data, realise repeatability conditions, and the stages realise intermediate precision conditions of measurement (VIM~2.22), in which device, surveyor and epoch are varied one axis at a time; only the site replication, where all operating factors change, approaches the reproducibility conditions of ISO~5725 \cite{ISO_Standard_2025_AccuracyTruenessAndPrecision}.

The site change between the two mines replicates the training stages of the scheme under a different magnetic distortion regime. The paper treats it as a replication of the protocol and therefore does not report it as a fifth stage.

At the cross-time stage three auxiliary quantities accompany \pmac{}. Where the class inventory has changed, precision is additionally reported conditioned on the classes still present. A neighbour-tolerant variant counts confusions with directly adjacent location segments as correct and thereby separates the ambiguity of the annotation boundaries from genuine signature change. The out-of-present-scope (OOPS) rate is the share of predictions assigned to classes that are no longer present in deployment relative to training; it exposes silent misdirection that a present-conditioned score alone would hide.

\subsection{Reference levels, acceptance criteria and decision rule}
\label{sec:protocol:criteria}

Table~\ref{tab:acceptance} states the acceptance criteria per stage. The reference levels use the correct class count of each stage: chance is $1/95$ in frame A, $1/40$ in frame B and $1/42$ for the present-conditioned cross-time metrics. The OOPS rate has a self-reference. A chance predictor over the 95 trained classes places $53/95=0.558$ of its predictions on classes absent at the re-measurement, and the declared ceiling is half this value, 0.279. This decision rule follows the guarded acceptance concept of JCGM~106 \cite{JointCommitteeforGuidesinMetrology_2012_EvaluationOfMeasurementData}: the acceptance limit is offset from the tolerance limit by a guard band, realised here by placing the OOPS ceiling at half the chance-level self-reference. The decision rule is quantile based: roll-out requires that the 5\,\% quantile of the repeat ensemble of the candidate configuration exceeds $k$ times chance and that, at the cross-time stage, the 95\,\% quantile of the OOPS rate stays below the ceiling. The JCGM~106 guard band applies to the OOPS ceiling alone. The quantile criterion compares the point estimate $\hat q_{0.05}$ with the threshold and is to that extent unguarded; section~\ref{sec:uncertainty} states from which ensemble size a non-parametric lower bound for $q_{0.05}$ exists at all, and reports it for the candidate configurations. The factor $k$ encodes the application. The class-exact metric governs rescue-zone identification and the neighbour-tolerant metric governs the navigation display.

\begin{table}
\caption{\label{tab:acceptance}Acceptance criteria per protocol stage. $K$ is the number of trained classes in the respective evaluation frame (mine A: $K=95$, mine B: $K=40$); at the cross-time stage the class-exact and neighbour-tolerant metrics are conditioned on the $K_{\mathrm{p}}=42$ classes present in the recordings of the re-measurement. $q_{0.05}$ denotes the 5\,\% quantile of the repeat ensemble of the candidate configuration. The factor $k$ is a deployment choice; verdicts in table~\ref{tab:verdicts} are evaluated at $k=5$ and $k=10$.}
\centering\small
\begin{tabular}{@{}llll}
\toprule
Stage & Reference level & Quantile criterion & OOPS criterion \\
\midrule
Random split & upper bound by declaration; & none (no roll-out & not applicable \\
 & window overlap reported & decision at this stage) & \\
Cross-device & chance $1/K$ & $q_{0.05} > k/K$ & structurally zero \\
Cross-subject & chance $1/K$ & $q_{0.05} > k/K$ & structurally zero \\
Cross-time & chance $1/K_{\mathrm{p}}$ (class-exact); & $q_{0.05} > k/K_{\mathrm{p}}$ & $q_{0.95}(\mathrm{OOPS}) <$ \\
 & tolerated-set prevalence $0.162$ & (class-exact; neighbour- & $0.5 \times 0.558 = 0.279$ \\
 & (neighbour, descriptive); & tolerant: descriptive, & \\
 & OOPS self-reference $53/95 = 0.558$ & no $k$-verdict) & \\
\bottomrule
\end{tabular}

\par\smallskip\noindent\footnotesize Decision rule: roll-out only if the quantile criterion and, at the cross-time stage, the OOPS criterion hold, with the class-exact metric for rescue zones and the neighbour-tolerant metric for the navigation display.

\end{table}

\subsection{Models, representations and configuration}
\label{sec:protocol:models}

The classifier under test is an LSTM \cite{Hochreiter_Schmidhuber_1997_LongShortTermMemory} over windows of the magnetic time series; the window length and step size are hyperparameters of the training campaigns. Eight input representations of the same measured field vector are compared throughout: the scalar total field $F_A=|\mathbf{B}|$; the Cartesian components $F_V=(B_x,B_y,B_z)$ in the device frame; a gravity-referenced parametrisation $F_K$ combining $|\mathbf{B}|$ with the angles between field and gravity vector; and the horizontal--vertical decomposition $F_{HZ}=(H,Z,\theta,I)$ in the world frame. Each set also exists in an attitude-augmented variant (suffix $+L$) that appends the device attitude (roll, pitch, yaw). The sets differ in what they preserve of the measured vector. $F_V$ and $F_{HZ}$ contain it in full, one in the device frame and one in the world frame; they are related by a rotation. $F_K$ keeps magnitude and inclination against gravity and discards the azimuth about the gravity axis. $F_A$ keeps only the magnitude and serves as a negative control. Where two sets carry the same information content and still behave differently under shift, that is a finding about the evaluation and not about the physics (section~\ref{sec:representations}).

Hyperparameter selection inside the model clouds evaluated here used a surrogate-based, attribution-guided procedure \cite{platte2026companion}. The present paper takes the clouds as given and asks what their members are worth under distribution shift. Its 7003 runs cover the six clouds of the two campaigns studied here together with their 1288 repeat runs, which carry the uncertainty statement of section~\ref{sec:uncertainty}. Every training run is recorded with its full configuration, all evaluation metrics and per-run timing under a deterministic configuration hash, which is what makes the retrospective application of the protocol in sections~\ref{sec:results1}--\ref{sec:uncertainty} reproducible.

\subsection{Training, evaluation and statistical procedures}
\label{sec:protocol:training}

Training used \texttt{CrossEntropyLoss} over at most 400 epochs with early stopping after ten unsuccessful epochs. In the training campaign this ended 99.4\,\% of the runs early, at a median of 74 epochs, and at 31 epochs in the uranium mine. Hyperparameters are the learning rate (\numlist{0.0001;0.001;0.01}), the learning-rate schedule (static or \texttt{ReduceOnPlateau}), the batch size (32 to 1024), the number and width of the LSTM layers (1 to 4 layers, 128 to 4096 units), and the window length (50 to 700 samples) and step size (1 to 100). The runs were distributed over two GPUs of one compute node.

Three items that a complete reproduction would require are absent from the archived training logs: the optimiser, the normalisation of the input quantities, and a random seed. The repeat ensembles of this work are therefore repetitions without a fixed seed. For their purpose, making the spread of the training process visible, that is precisely the required condition; for a bit-exact reproduction of individual runs the logging is not sufficient.

One peculiarity of the archived evaluation must be disclosed. Two evaluation notebooks multiplied the macro-averaged precision and recall columns of the $F_{A+L}$ rows by constant factors before saving: 0.8 for the 63 cross-subject runs of the training campaign and 0.7 for all runs of the re-measurement. The notebooks give no reason for the scaling. All $F_{A+L}$ values reported here are descaled. For the re-measurement the descaling was confirmed independently from the stored confusion matrices (largest deviation below $10^{-12}$), and for the training campaign through the copies of the same models carried in the re-measurement logs. The OOPS rate, the confusion matrices and the per-class vectors were not scaled.

The following holds throughout for the statistical statements. Bootstrap intervals are percentile intervals from \num{20000} draws with seed 42. Paired comparisons report the share of positive pairs alongside the Wilcoxon signed-rank test, because the difference distributions are markedly asymmetric and the Wilcoxon test then makes no clean statement about the median. All $p$ values are uncorrected; where several tests jointly carry a claim, we additionally give the corrected decision.

\section{Results: reference levels and the first three stages}\label{sec:results1}

\subsection{Reference levels: the random split as a declared upper bound}
\label{sec:optimism_gap}

\begin{figure}
\centering
\includegraphics[width=\linewidth]{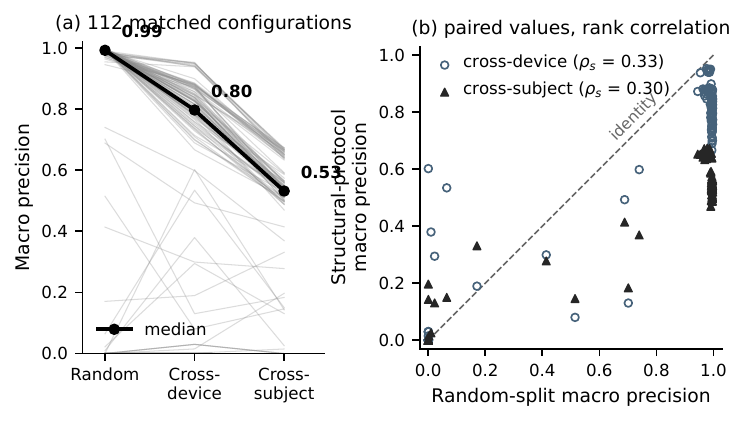}
\caption{\label{fig:protocol_gap}Optimism gap of the random split in evaluation frame A. (a) Macro precision of 112 matched configurations under the three protocols; thin lines connect the values of one configuration, the black line connects the protocol medians. (b) The same values paired against the random-split result; open circles mark the cross-device protocol, filled triangles the cross-subject protocol, the dashed line marks identity. Spearman rank correlations are annotated.}
\end{figure}

In evaluation frame A, 112 hyperparameter configurations were trained under all three splitting protocols. The members of each triple are identical in every varied hyperparameter (feature set, sequence length, window step size, LSTM width and depth, batch size, learning rate); the learning-rate schedule remains as a residual difference between the campaigns (static in the randomised screening, adaptive in the structural protocols). A within-campaign sensitivity check bounds this confound: in frame~B, where both schedules occur in the random screening, 35 configuration pairs matched in all other hyperparameters show a median advantage of $+0.007$ for the adaptive schedule (interquartile range 0.005--0.014, 31 of 35 pairs positive, Wilcoxon $p=1.1\times10^{-6}$), an order of magnitude below the gaps reported in this section. It does not hold for the far smaller ageing differences in section~\ref{sec:stage4-longtime}, where the residual difference exceeds the effect. Figure~\ref{fig:protocol_gap}(a) shows the paired values. The median macro precision is 0.992 under the random split, 0.797 under the cross-device protocol and 0.532 under the cross-subject protocol. Paired per configuration, the random split exceeds the cross-device result by a median of $+0.143$ (95\,\% bootstrap confidence interval $[0.114, 0.163]$, Wilcoxon $p<10^{-12}$, 80\,\% of pairs positive) and the cross-subject result by $+0.413$ ($[0.336, 0.443]$, $p<10^{-17}$, 94\,\% of pairs positive).

The random split also carries little information for model selection. Over all 112 triples the Spearman rank correlation between the random-split ranking and the deployment rankings is $\rho_s=0.33$ ($p=3\times10^{-4}$) for cross-device and $\rho_s=0.30$ ($p=1.5\times10^{-3}$) for cross-subject, while the two structural protocols correlate with each other at $\rho_s=0.89$ (figure~\ref{fig:protocol_gap}(b)). Those two weakly positive values, however, come entirely from the 16 configurations of the negative control $F_A$, which sit near zero under all three protocols and thereby create a two-point correlation between \enquote{degenerate} and \enquote{not degenerate}. Over the remaining 96 configurations the association vanishes ($\rho_s=-0.05$ and $-0.11$, $p>0.25$); restricted to the 83 configurations that converge above 0.95 in the random split it reverses ($\rho_s=-0.56$ and $-0.65$). The agreement between the two structural protocols persists ($\rho_s=0.82$ without $F_A$). The optimism gap itself grows on these subsets ($+0.160$ and $+0.438$ without $F_A$). The protocol therefore assigns the random split a single role: it is reported as the declared upper bound of the achievable precision on the given data.

\subsection{Leakage mechanism and reporting rule}
\label{sec:leakage}

Consecutive windows generated with step size $s$ from a time series share, at window length $L$, exactly $(L-s)/L$ of their samples. At the operating point $L=700$, $s=1$ the overlap is 99.86\,\%, and a random split then places near-duplicates of most test windows into the training set. Within the 1306 random-split runs of frame A, macro precision increases with window overlap (Spearman $\rho_s=0.39$, $p<10^{-47}$); frame B replicates the direction on its narrower grid of overlaps 0.990 to 0.9986 ($\rho_s=0.18$, $n=1185$, $p=2\times10^{-10}$). Configurations matched in all other hyperparameters isolate the gradient: raising the step size from 2 to 20 costs the random split a median of 0.042 macro precision (43 pairs), and raising it from 2 to 100 costs 0.256 (14 pairs). Under the structural protocols the available manipulation from $s=1$ to $s=2$ shifts the median by at most 0.011 (cross-device $+0.009$, 192 pairs; cross-subject $+0.010$, 248 pairs; random $+0.002$, 20 pairs), as both settings keep the overlap above 99.7\,\%. The step-size gradient combines two effects, removal of near-duplicate test windows and shrinkage of the training set, and the small structural-protocol differences bound the shrinkage effect for a factor-two step increase. Under structural splitting the leakage path is closed by construction because no window pair can span the device or subject boundary. The resulting reporting rule is part of the protocol: every random-split figure is accompanied by the window overlap $(L-s)/L$, and random-split results enter the evaluation solely as the declared upper bound.

\subsection{Stages 1--3: random split, device change, subject change}\label{sec:stages}

\begin{figure}
\centering
\includegraphics[width=\linewidth]{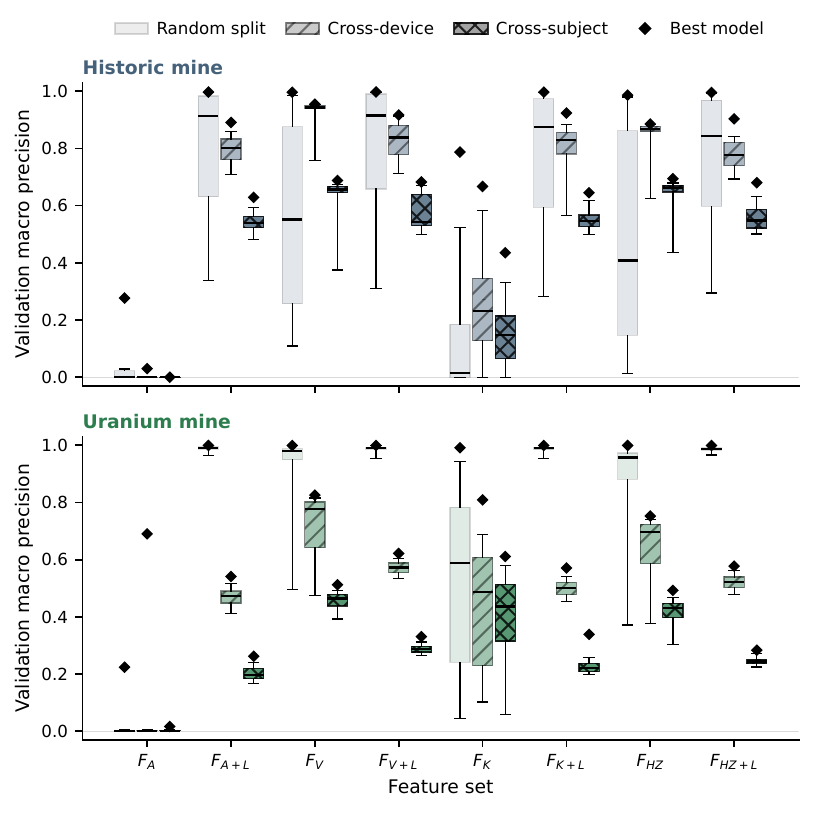}
\caption{Validation macro precision of all trained configurations per feature set and evaluation stage in the historic mine (top, $n=1306/416/504$ runs) and the uranium mine (bottom, $n=1185/1152/1152$ runs). Boxes give median and interquartile range, whiskers the 5th and 95th percentile, diamonds the best model per feature set and stage. Stages are distinguished by fill tone and hatching.}
\label{fig:escalation}
\end{figure}

Figure~\ref{fig:escalation} summarises the first three stages of the protocol in both mines. Each box aggregates all trained configurations of one feature set at one stage. All distribution statements use quantiles (median, interquartile range, whiskers at the 5th and 95th percentile) because the configuration distributions are strongly skewed and contain divergent runs. The best model per feature set and stage is marked separately, since the protocol declares the random-split optimum as an upper bound rather than an expected operating point. Table~\ref{tab:best_models} lists the best model per stage and mine.

In the historic mine the attainable optimum degrades from 0.9974 in the random split ($F_{V+L}$) to 0.9544 under device change ($F_V$) and to 0.6942 under subject change ($F_{HZ}$, $n=504$ runs at this stage). The uranium mine replicates the pattern at lower levels under shift: 0.9993 ($F_V$) in the random split, 0.8258 ($F_V$) under device change, 0.6111 ($F_K$) under subject change ($n=1152$ runs per shift stage). For the most transfer-stable vector representation $F_V$ the cross-subject optimum in the uranium mine is 0.5126 with a median of 0.4646 over 144 configurations. The shift-stage distributions are narrow for the informative feature sets, for example $F_{HZ}$ under subject change in the historic mine with a median of 0.6618 and an interquartile range of 0.6478 to 0.6688. The reported optima are representative of their configuration neighbourhood in the sense of these narrow distributions, with one instructive exception: the cross-subject optimum of the historic mine (configuration \texttt{271e769f83}) stems from a bimodal training process and is analysed as such in section~\ref{sec:oneshot-risk}. The random stage of the historic campaign doubled as a hyperparameter screening over a broad, randomly sampled search space (1306 runs) and consequently contains many divergent configurations. Its medians characterise that sample, and comparisons across stages therefore rely on the attainable optima and on the per-stage distributions shown in figure~\ref{fig:escalation}.

\begin{table}
\caption{Best model per evaluation stage and mine (validation macro precision, 95 classes in the historic mine, 40 classes in the uranium mine). Model ids abbreviate the configuration hash of the training campaign.}
\label{tab:best_models}
\centering
\begin{tabular}{llllr}
\toprule
Mine & Stage & Feature set & Model id & \pmac \\
\midrule
Historic & Random split & $F_{V+L}$ & \texttt{3793ef462f} & 0.9974 \\
Historic & Cross-device & $F_V$ & \texttt{0f9393c1a7} & 0.9544 \\
Historic & Cross-subject & $F_{HZ}$ & \texttt{271e769f83} & 0.6942$^{\rm c}$ \\
\midrule
Uranium & Random split & $F_V$ & \texttt{22ea54a20b} & 0.9993 \\
Uranium & Cross-device & $F_V$ & \texttt{5f23fe76a7} & 0.8258 \\
Uranium & Cross-subject & $F_K$ & \texttt{bcb8d9012f} & 0.6111 \\
\bottomrule
\end{tabular}

\smallskip\noindent{\footnotesize $^{\rm c}$ stems from a bimodal training process; the repeat
ensemble of this configuration has median 0.943 and the tabulated campaign run sits in a rare low mode (section~\ref{sec:oneshot-risk}).}
\end{table}

\subsection{Representation findings}\label{sec:representations}

\paragraph{Failure of the scalar total field.}
The total-field magnitude $F_A$ fails at every stage that involves a distribution shift. Under subject change in the historic mine no configuration exceeds a macro precision of 0.0001 (63 runs), and in the uranium mine the best of 144 configurations reaches 0.0167. Under device change the historic-mine optimum is 0.0302. In the uranium mine 95\% of the cross-device runs stay below 0.008, with a single configuration reaching 0.6903. A scalar field magnitude passes the random split in combination with attitude channels (see below) and is unusable as soon as the device or the person changes.

\begin{figure}
\centering
\includegraphics[width=\linewidth]{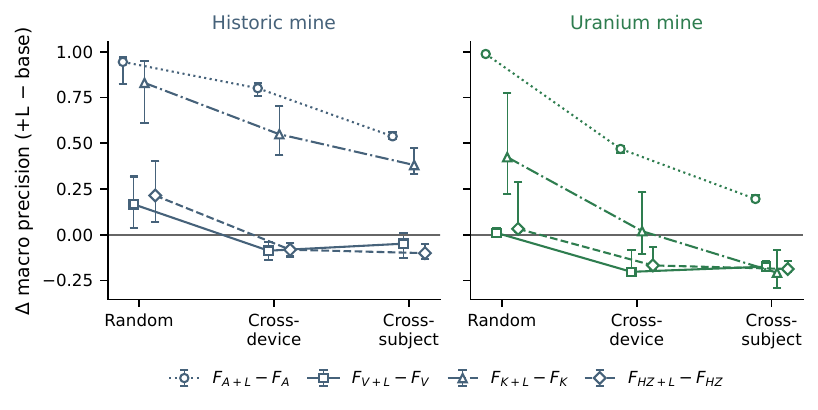}
\caption{Paired differences in validation macro precision between each attitude-augmented feature set and its base set, matched on identical remaining hyperparameters within each stage and mine. Markers give the median of the paired differences, bars the interquartile range. Pairs are distinguished by marker shape and line style.}
\label{fig:representation_pairs}
\end{figure}

\paragraph{Attitude channels help in the random split and hurt under shift.}
To isolate the effect of the attitude channels (roll, pitch, yaw) we paired each augmented feature set with its base set on identical values of the nine remaining hyperparameters within each stage and mine. Figure~\ref{fig:representation_pairs} shows the paired median differences. For the informative vector representations the pattern is consistent in both mines. The $F_{V+L}$ versus $F_V$ difference in the uranium mine moves from $+0.010$ in the random split to $-0.202$ under device change and $-0.174$ under subject change, where 0 of 144 pairs are positive. The corresponding values in the historic mine are $+0.167$, $-0.087$ and $-0.049$. The $F_{HZ}$ pair behaves in the same way ($+0.215$, $-0.082$, $-0.100$ historic; $+0.032$, $-0.167$, $-0.187$ uranium). All 24 paired differences are significant at $p<0.01$ (two-sided Wilcoxon signed-rank test); 22 of them remain so after Bonferroni correction over the 24 tests. The two exceptions are $F_{V+L}$ against $F_V$ under subject change in the historic mine and $F_{K+L}$ against $F_K$ under device change in the uranium mine. The latter is undetermined in sign anyway: its median is $+0.020$ at an interquartile range of $-0.105$ to $+0.233$ with 52.8\,\% positive pairs. For the impoverished representations the attitude channels act as a substitute signal instead: the $F_A$ and $F_K$ pairs remain positive under shift in the historic mine ($+0.538$ and $+0.381$ under subject change), and only the uranium mine turns the $F_K$ pair negative under subject change ($-0.207$).

This scissors effect shifts the ranking of the representations between stages. In the random split the four attitude-augmented sets occupy the top four median ranks in both mines. Under device change $F_V$ ranks first in both mines, and under subject change the attitude-free base sets lead ($F_{HZ}$ before $F_V$ in the historic mine, $F_V$ before $F_K$ and $F_{HZ}$ in the uranium mine). The Kendall rank correlation between the random-split ranking and the rankings of the shift stages over the eight sets is indistinguishable from zero in all four comparisons: against cross-device $\tau=0.36$ ($p=0.28$) in the historic mine and $\tau=0.00$ ($p=1.00$) in the uranium mine, against cross-subject $\tau=0.00$ ($p=1.00$) and $\tau=-0.29$ ($p=0.40$). What these four values can establish is limited by the sample size. With eight feature sets the smallest attainable significant magnitude is $|\tau|=18/28=0.64$ at an exact two-sided $p$ of 0.031. Against that rejection region the test has a power of 0.14 against a true agreement of $\tau=0.33$, of 0.35 against $\tau=0.5$ and of 0.75 against $\tau=0.7$. The four non-significant values therefore rule out strong rank agreement, but not moderate agreement. What is defensible is the point estimate: at $\tau\le0.36$ it lies well below the agreement the two shift stages reach with each other ($\tau=0.64$, $p=0.031$ historic; $\tau=0.71$, $p=0.014$ uranium). The two mines also agree on which representations transfer, with cross-mine values of $\tau=0.86$ ($p=0.002$) in the random split, $\tau=0.79$ ($p=0.006$) under device change and $\tau=0.43$ ($p=0.18$) under subject change. All $p$ values are exact permutation values over the $8!$ rankings. Two qualifications belong with this. First, $F_A$ occupies rank 8 in all six stages and thereby contributes 7 of the 28 pairs concordantly by construction; over the seven sets that did learn, the two agreements between the shift stages fall to $\tau=0.52$ ($p=0.14$) and $\tau=0.62$ ($p=0.069$) and are then no longer significant, while the cross-mine agreement in the random split persists at $\tau=0.81$ ($p=0.011$). Second, none of the nine tests in this paragraph survives Holm correction at $\alpha=0.05$. What remains as a secured individual finding is the cross-mine stability of the ranking, not the agreement of the shift stages with each other.

\paragraph{Same information, different parametrisation.}
The Cartesian components $F_V$ and the horizontal--vertical decomposition $F_{HZ}$ encode the same magnetic vector. Their optima drop by similar amounts from random split to subject change ($-0.309$ for $F_V$ and $-0.292$ for $F_{HZ}$ in the historic mine, $-0.487$ and $-0.507$ in the uranium mine). Under device change $F_V$ holds an advantage in both mines, with medians of 0.9435 versus 0.8688 in the historic mine and 0.7782 versus 0.6976 in the uranium mine. Under subject change the two parametrisations converge: $F_{HZ}$ is nominally ahead in the historic mine (median 0.6618 versus 0.6563, best model 0.6942 versus 0.6879), a difference within the configuration spread of both sets, and $F_V$ ahead in the uranium mine (median 0.4646 versus 0.4317). The choice among mathematically equivalent parametrisations of the same measured vector shifts the cross-device median by up to 0.08 and is itself a protocol-relevant design decision.

\section{Stage 4: cross-time evaluation after 34 months}
\label{sec:stage4-longtime}

Stage~4 re-evaluates the unchanged models of the training campaign on data recorded 34 months after the training recordings. The recordings of the re-measurement were made with a smartphone model unseen in training (iPhone~16~Pro). In the Cross-Subject cell the recording subject was also unseen, so this cell accumulates all three shift axes of the protocol: unseen device, unseen subject, and 34 months of elapsed time. Of the 95 location classes of the training campaign, 42 were still accessible and annotated at the re-measurement. Present-conditioned metrics refer to these 42 classes, with a chance level of $1/42 \approx 0.024$. The neighbour-tolerant score uses the convention of the archived pipeline and averages over the classes that the model actually predicts, whereas the strict score counts never-predicted classes as zero (equation~\ref{eq:pmac}); the decomposition states both conventions explicitly where they meet. Predictions assigned to one of the 53 classes no longer present in deployment are counted by the out-of-present-scope (OOPS) rate. By \emph{cell} we mean in the following the set of all runs a feature set occupies under one training protocol. Each cell is summarised by the median and the interquartile range (IQR) across its training runs ($n=63$ per feature set in the Cross-Subject cell).

\paragraph{Pipeline integrity.}
Before interpreting long-term scores we checked the reconstructed evaluation pipeline. For 79 archived models from two configurations ($F_V$: 71 runs, $F_{HZ}$: 8 runs) the validation data of the training campaign were re-scored with the pipeline of the re-measurement. The re-computed values lie systematically $1.1$~percentage points (pp) below the archived values of the training campaign, with a largest absolute deviation of $1.2$\,pp and a Pearson correlation of $r=0.999$. This is an offset, not scatter. The control covers two configurations; for the remaining feature sets, in particular the unstable $F_K$ family, it is not replicated. Every long-term difference reported below therefore carries a declared systematic contribution of up to $1.2$\,pp.

\begin{table}
\caption{Stage 4, cumulative Cross-Subject cell: models of the training campaign, evaluated on the data of the re-measurement (device unseen in training, subject unseen in training, no retraining). Medians across $n=63$ runs per feature set, IQR in brackets. 95 trained classes, 42 present at the re-measurement, chance level $1/42\approx0.024$. $F_A$ did not learn in the training campaign and serves as a negative control.}
\label{tab:stage4-longtime}
\centering
\small
\begin{tabular}{lccccc}
\toprule
 & Val.\ training & \multicolumn{2}{c}{Present-cond.\ precision} & Neighbour-tol. & OOPS rate \\
Set & (strict) & median & IQR & median & median \\
\midrule
$F_A$      & 0.000 & 0.001 & [0.001, 0.001] & 0.155 & 0.000 \\
$F_{A+L}$  & 0.538 & 0.334 & [0.323, 0.350] & 0.580 & 0.108 \\
$F_V$      & 0.656 & 0.422 & [0.409, 0.428] & 0.636 & 0.184 \\
$F_{V+L}$  & 0.544 & 0.340 & [0.330, 0.349] & 0.598 & 0.116 \\
$F_K$      & 0.149 & 0.187 & [0.077, 0.237] & 0.374 & 0.434 \\
$F_{K+L}$  & 0.547 & 0.355 & [0.341, 0.366] & 0.559 & 0.415 \\
$F_{HZ}$   & 0.662 & 0.356 & [0.346, 0.371] & 0.521 & 0.164 \\
$F_{HZ+L}$ & 0.548 & 0.339 & [0.328, 0.350] & 0.576 & 0.095 \\
\bottomrule
\end{tabular}
\end{table}

\paragraph{Cumulative result.}
Table~\ref{tab:stage4-longtime} summarises the cell with the strongest cumulative shift. The best feature set $F_V$ reaches a median present-conditioned macro precision of 0.422 (IQR 0.409--0.428), a factor of 17.7 above chance. With one-neighbour tolerance the median rises to 0.636 (IQR 0.621--0.645). The training regime is associated with ageing, and the size of the association depends on how it is measured. In cell medians the long-term $F_V$ scores order by training regime (0.319 random, 0.411 cross-device, 0.422 cross-subject), but the cells differ in composition, exactly the comparison section~\ref{sec:stages} warns against. Matched on the 112 configuration triples the ordering persists with far smaller margins (0.333, 0.342, 0.343). The margin of cross-device against random is $+0.004$ (95\,\% bootstrap confidence interval of the median $[0.000, 0.009]$, Wilcoxon $p=0.0013$, sign test 62 of 112 positive, $p=0.30$), and of cross-subject against random $+0.0035$ ($[0.0002, 0.0128]$, $p=3.5\times10^{-5}$, 66 of 112, $p=0.072$); cross-subject against cross-device is not significant. Two reasons forbid reading a genuine advantage of the harder regimes into this. The confidence interval of the first comparison includes zero, and the distribution-free sign test is non-significant for both. Above all, the effect lies below the residual learning-rate-schedule difference that section~\ref{sec:optimism_gap} quantifies at $+0.007$ in favour of the adaptive schedule and that acts in the same direction. What remains defensible is the null statement: models trained under harder validation regimes age no faster. The cell-median gap of ten points does not measure ageing but the differing composition of the cells. For the ten matched $F_V$ triples the margin is larger (0.396, 0.408, 0.425, all ten pairs positive); the learning-rate schedule co-varies there as well.

\begin{figure}
\centering
\includegraphics[width=\columnwidth]{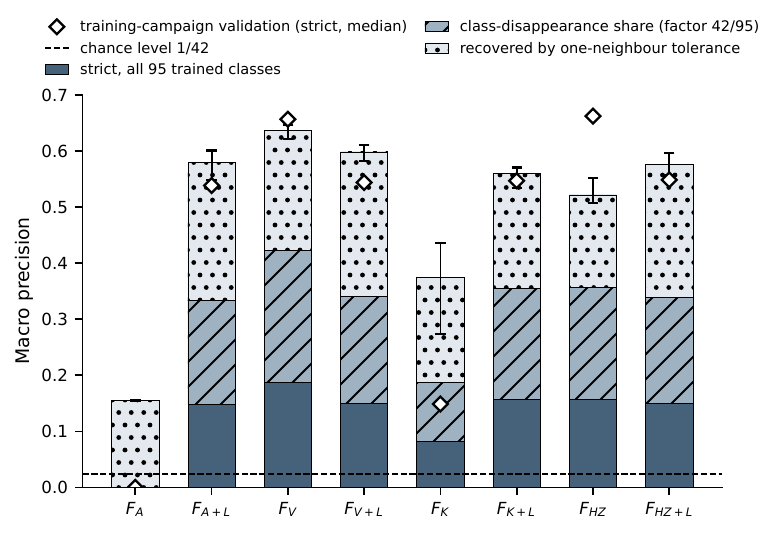}
\caption{Decomposition of the long-term degradation in the cumulative Cross-Subject cell (medians over $n=63$ runs per feature set). The bars stack the strict all-classes score, the share removed by class loss (exact factor $42/95$), and the share recovered by one-neighbour tolerance. Diamonds mark the strict validation median of the training campaign. Whiskers give the IQR of the neighbour-tolerant score. The dashed line marks the chance level $1/42$.}
\label{fig:longtime-decomposition}
\end{figure}

\begin{figure}
\centering
\includegraphics[width=\columnwidth]{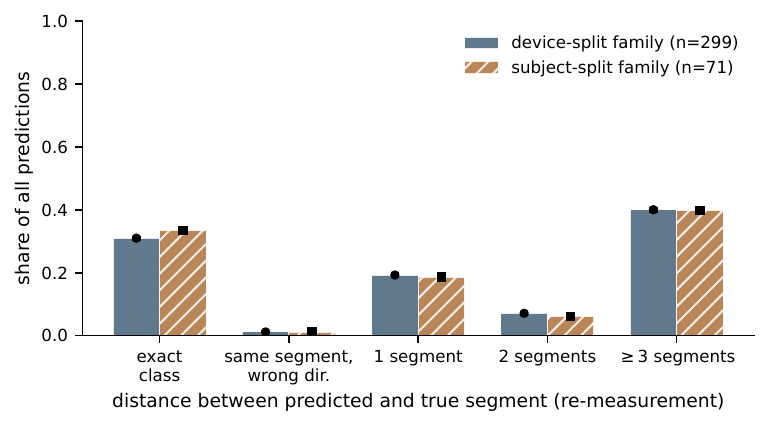}
\caption{Topological error distance of misclassified windows in the cumulative Cross-Subject cell (repeat ensembles of the two stage-4 candidate configurations; breadth-first-search distance on the gallery adjacency graph). Recovered windows lie in adjacent segments; the tail of the distribution reaches distant segments.}
\label{fig:error-distance}
\end{figure}

\paragraph{Decomposition of the apparent degradation.}
A naive long-term reading scores all 95 trained classes strictly. For $F_V$ this yields a median of 0.187, which is 47.0\,pp below the training-campaign validation median of 0.656. Figure~\ref{fig:longtime-decomposition} decomposes this drop into three effects of different origin. (i)~Class loss: with 53 of 95 classes absent, macro precision over all trained classes equals the present-conditioned value multiplied by $42/95$. The archived per-run values reproduce this identity to numerical precision (largest deviation $3\times10^{-16}$). For $F_V$ this accounts for 23.5\,pp of the gap to the present-conditioned value and is a property of the metric and the class inventory. (ii)~Boundary tolerance: counting confusions with directly adjacent location segments as correct recovers 16.5--25.8\,pp across the seven feature sets that learned at all (21.4\,pp for $F_V$). The negative control $F_A$ is excluded here: it predicts a single class only, so its neighbour-tolerant value of 0.155 is the value of that one class under the averaging convention of the archived pipeline and cannot carry a reference level (section~\ref{sec:oneshot-risk}). The error-distance distribution (figure~\ref{fig:error-distance}) qualifies this recovery. Within the repeat ensembles of the two stage-4 candidate configurations, 51--53\% of all windows fall within one segment of the true location, but among the misclassified windows just under 28\% lie in an adjacent segment (1.6\% in the correct segment with the wrong direction). A uniform random predictor already reaches 14.2\% there, since on average 5.8 of the 41 remaining present classes lie in an adjacent segment, so about half of the tolerance gain is what guessing produces as well. The median error distance is 4 segments and the 95th percentile exceeds 20. The neighbour tolerance therefore recovers the numerous windows near annotated boundaries, and the residual errors include long-range confusions rather than only local ambiguity. (iii)~Residual: the remaining gap to the training-campaign level is 2.1\,pp for $F_V$ and 14.0\,pp for $F_{HZ}$. For the five other learned feature sets the residual is negative, because the training-campaign reference is a strict score without neighbour tolerance and the neighbour-tolerant score of the re-measurement exceeds it. The genuine change of the magnetic signatures for $F_V$ is therefore bounded between 2.1\,pp (neighbour-tolerant reading) and 23.4\,pp (strict present-conditioned reading). The lower bound presupposes that the entire gain of the neighbour tolerance is due to blurred annotation boundaries, and it moreover compares a neighbour-tolerant value with a strict reference. Both are optimistic. What is defensible is the upper bound.

\begin{figure}
\centering
\includegraphics[width=\columnwidth]{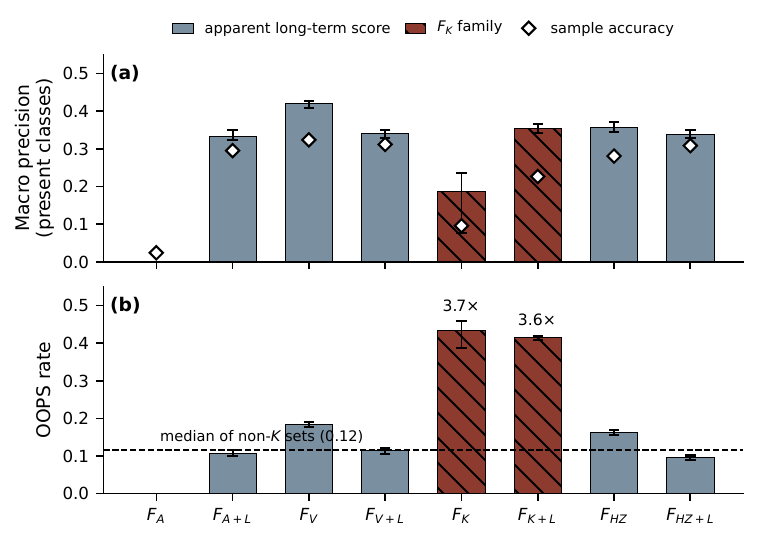}
\caption{Apparent performance and OOPS rate in the cumulative Cross-Subject cell (medians, whiskers: IQR). (a)~Present-conditioned macro precision (bars) and sample accuracy (diamonds). (b)~OOPS rate, the share of predictions assigned to the 53 classes absent at the re-measurement. The $F_K$ family combines competitive present-conditioned scores with OOPS rates 3.6 to 3.7 times the median of the other learned feature sets.}
\label{fig:oops}
\end{figure}

\paragraph{The OOPS artefact.}
Present-conditioned precision ignores predictions on classes no longer present in deployment, since such predictions enter no precision denominator of a present class. Figure~\ref{fig:oops} shows the consequence. $F_{K+L}$ attains a median present-conditioned precision of 0.355 and thus even sits slightly above $F_{V+L}$ (0.340) and $F_{HZ+L}$ (0.339). At the same time $F_K$ and $F_{K+L}$ direct 43\% and 41\% of their predictions to classes that no longer exist at the re-measurement (OOPS medians 0.434 and 0.415, against 0.095--0.184 for the other learned feature sets, factors of 3.7 and 3.6 relative to their median of 0.116). Sample accuracy exposes the loss: 0.227 for $F_{K+L}$ against 0.312 for $F_{V+L}$ and 0.309 for $F_{HZ+L}$. A present-conditioned score is interpretable only together with the OOPS rate, which the protocol therefore reports as a mandatory plausibility metric.

\paragraph{Run-to-run uncertainty.}
The repeat-run cell of the long-term frame contains 1031 re-evaluations of runs with identical configurations. For the stable configurations the run-to-run IQR of present-conditioned precision is 0.012--0.015 ($F_V$: medians 0.407 and 0.430, $F_{HZ}$: 0.353 and 0.370). The $F_K$ configuration is bimodal: its 77 runs span 0.001 to 0.434 with an IQR of 0.379, and a fraction of the runs collapses to chance level during training. This instability also explains the near-zero $F_K$ median in the random-split cell. A mean over such a cell mixes collapsed and functional runs into a value that no single run attains. All aggregates in this work are therefore quantiles.

\section{Uncertainty of the deployed model: repeated trainings and verdicts}\label{sec:uncertainty}

\subsection{One-shot risk assessment}
\label{sec:oneshot-risk}

A deployed model is the outcome of a single training run. Repeating an identical configuration on identical data therefore measures the risk that this one draw from the training process falls short of the reported performance. Run-to-run variation stems from training stochasticity alone, that is, from weight initialisation, minibatch shuffling and nondeterministic GPU reductions. The repeat ensembles comprise five configurations in the historic mine (71 to 299 de-duplicated runs each), the same models re-evaluated on the data of the re-measurement without retraining, three configurations in the uranium mine (9 to 78 runs). One further configuration with only three repeats was excluded.

The reference for all stages is the chance level of macro-averaged precision under uniform random prediction. The expected precision of each class then equals its prevalence, and prevalences average to $1/K$ across the $K$ classes. The chance level is therefore $1/K$ regardless of class imbalance, with $K=95$, $42$ and $40$ in the three frames.

The assessment proceeds in three steps per configuration. Step~1 tests the mean of the repeat distribution against chance with a one-sided one-sample $t$-test; for the ensembles with $n\ge71$ the central limit theorem covers the strongly non-normal shapes reported below; for $n=24$ and $n=9$ it does not, and there the distribution-free sign test is the operative figure. An exact sign test of whether a single run exceeds chance with probability above 0.5 serves as a distribution-free control. We avoid the Wilcoxon signed-rank test because its symmetry assumption fails for bimodal distributions. Step~2 reports the share of above-chance runs together with the lower bound of its two-sided 99\,\% Clopper--Pearson interval, which is the success share an operator can evidence from the campaign. Step~3 reports the empirical 5\,\% quantile $q_{0.05}$ (linear interpolation between order statistics) as the conservative minimum performance of a single deployment, expressed as a factor over chance. At the available campaign sizes the 1\,\% quantile would rest on the single weakest run. The order statistic behind the empirical 5\,\% quantile depends on $n$: only for $n\ge59$ does $\mathrm{P}(\min > q_{0.05})=0.95^{n}$ fall below 0.05, for $n$ between about 24 and 59 it rests on the second to fourth order statistic, and for $n=9$ it interpolates between the weakest and the second-weakest run. More sharply still: for $n\le58$ no valid non-parametric 95\,\% lower bound for the 5\,\% quantile exists at all, the minimum included; only from $n=59$ does the minimum become such a bound. The minimal adoption of section~\ref{sec:cost} therefore sets $n=59$ as the \emph{lower bound for guarded quantile-based verdicts}. That costs 22.0 GPU\,h instead of 11.2, and thus 0.68\,\% instead of 0.35\,\% of the study reported here. A mere point estimate of the quantile gets by with $n=30$; $n=10$ is usable only as a screening size. For the verdicts of this work the distinction is without consequence. Every repeat ensemble that carries a verdict in table~\ref{tab:verdicts} comprises at least 71 runs, and its minimum, which at those sizes is the valid 95\,\% lower bound, exceeds even the threshold for $k=10$: 0.650 against 0.105 under subject change in the historic mine, 0.381 and 0.411 against 0.238 at the cross-time stage, and 0.633 against 0.250 in the uranium mine. All verdicts therefore hold in the guarded reading as well. The two small ensembles U2 ($n=24$) and U3 ($n=9$) fall below that size but carry no verdict in table~\ref{tab:verdicts}; they appear in table~\ref{tab:oneshot-risk} only.

\begin{figure}
\centering
\includegraphics[width=\linewidth]{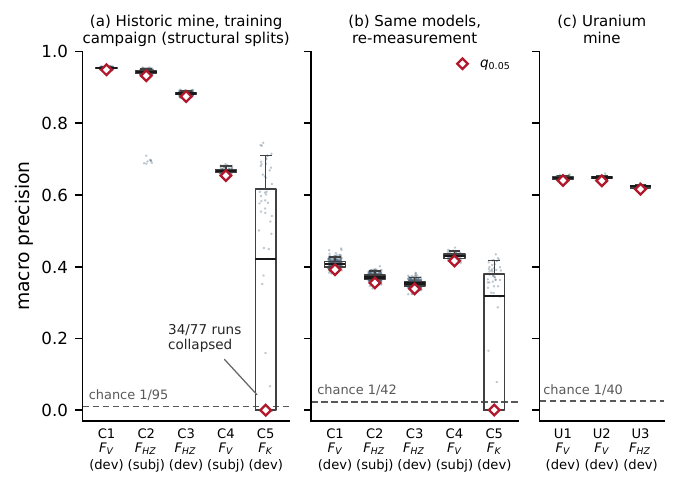}
\caption{repeated trainings of identical configurations. Points show individual runs, boxes the interquartile range with whiskers at the 5th and 95th percentiles, open diamonds the empirical 5\,\% quantile $q_{0.05}$, dashed lines the chance level $1/K$. (a)~Historic mine, training campaign, structural splits ($K=95$). (b)~The same models re-evaluated on the data of the re-measurement without retraining, macro precision conditioned on the classes present there ($K=42$). (c)~Uranium mine ($K=40$).}
\label{fig:oneshot-risk}
\end{figure}

Table~\ref{tab:oneshot-risk} and figure~\ref{fig:oneshot-risk} summarise the assessment. All thirteen ensemble evaluations pass step~1 with $p \le 10^{-12}$. Two qualifications belong with that. The thirteen evaluations cover eight configurations; five of them enter twice, once on the validation data of the training campaign and once on those of the re-measurement, and are to that extent not independent. And for U2 ($n=24$) and U3 ($n=9$) the operative figure is not the $t$ value but the exact sign test at $p=6.0\times10^{-8}$ and $p=2.0\times10^{-3}$; the latter is the smallest value attainable with nine runs. Step~3 separates them. In the structural splits of the training campaign the 5\,\% quantiles of the four stable configurations reach factors of 62 to 90 over chance. After 34 months the four stable configurations retain factors of 14.2 to 17.5 on the present-conditioned metric, so even the conservative lower bound of a single run remains an order of magnitude above chance. In the uranium mine the factors are 24.6 to 25.6.

The $F_K$ configuration C5 shows why step~1 cannot stand alone. Its repeat distribution splits into two groups: 34 of 77 runs collapse during training and all return exactly the same degenerate value of $3.24\times10^{-4}$ (predicting a single class, standard deviation within the group zero), the remaining 43 runs span 0.067 to 0.746, and the standard deviation over all runs is 0.31. Only 5.2\,\% of the runs fall in the transition band between 0.05 and 0.4. A Gaussian mixture is the wrong instrument here, because one of its components meets a point mass; the finding is sharper and exact without one. The $t$-test still signals $p = 9.5\times10^{-14}$ because the mean of 0.325 lies far above chance. The sign test drops to $p = 0.18$, the evidenced success share falls to 43/77 with a 99\,\% lower bound of 0.406, and $q_{0.05} = 3.2\times10^{-4}$ equals a factor of 0.03 over chance. For this configuration the measurand of configuration quality is the repeat distribution.

The verdicts are robust to the choice of macro precision as the measurand. We prefer it because a wrongly reported location could under some circumstances misdirect the user of the system in a mine. Macro recall of the same ensembles gives $q_{0.05}=0.948$ (C1), $0.651$ (C4), $0.415$ and $0.448$ (cross-time device-split and subject-split ensembles) and $0.615$ (U1); no verdict changes.

Two further observations concern the reporting itself. First, the Clopper--Pearson bound makes the cost of small repeat ensembles explicit. Step~2 states the share of above-chance runs through the lower bound of its two-sided 99\,\% confidence interval. If every single run of an ensemble is above chance, that bound depends on the size of the ensemble alone: with 9 out of 9 runs it evidences a share of 0.555, with 299 out of 299 runs one of 0.982. The size of the repeat ensemble therefore bounds the share that can be evidenced, independently of how well the runs perform. Second, the 5\,\% quantile has a blind spot for rare failure modes. Configuration C2 contains a minority mode of 8 out of 282 de-duplicated runs at precisions of 0.687 to 0.709 instead of the majority mode near 0.94. Its weight of 2.8\,\% lies below the quantile level, so $q_{0.05}=0.932$ does not expose it. This minority mode also resolves an apparent contradiction with table~\ref{tab:best_models}: the single cross-subject campaign run of the same configuration (0.694, the tabulated stage optimum) falls exactly into this rare low mode. Re-scored on the common validation data of the training campaign, the campaign model reaches 0.702, just above the four evaluable low-mode ensemble members, which reach 0.676 to 0.686 there, and far below the majority mode (0.980--0.983). The values 0.687 to 0.709 above come from the archived evaluation, the values 0.676 to 0.686 from this joint re-scoring; of the eight low-mode runs, four have no re-scored value. under the re-evaluation on the data of the re-measurement the mode difference disappears (the campaign model sits at the 45th percentile of its ensemble). The tabulated optimum is itself a weak draw of the training process, which is the strongest available argument for ensemble-based acceptance evidence. The converse limitation must be stated as well: the validation assignment of the C2 repeat batch is not reconstructable from the archived logs, and its quantile ($q_{0.05}=0.932$) exceeds every one of the 504 cross-subject screening runs. C2 therefore serves here as evidence about training stochasticity and must not be read as a cross-subject performance statement; the cross-subject verdicts of table~\ref{tab:verdicts} rest exclusively on C4. The share statistic of step~2 and the plotted distributions therefore remain part of the report alongside the quantile.

All runs of a campaign share one data split. The assessment quantifies robustness against training stochasticity and makes no population inference beyond it.

\begin{table}
\caption{Three-step one-shot risk assessment of the repeat trainings. Each row is one hyperparameter configuration trained $n$ times. The measurand is macro-averaged precision (block (b): conditioned on the classes present at the re-measurement). Split names the held-out unit of the train--test split as recorded in the archived training logs (device: phone-based, subject: person-based). Chance level is $1/K$. $q_{0.05}$ is the empirical 5\,\% quantile (linear interpolation), $q_{0.05}/\mathrm{chance}$ the factor over chance. $p$ is the one-sided one-sample $t$-test of the mean against chance (exact value per row). $\hat{p}_{>\mathrm{chance}}$ is the share of runs above chance together with the lower bound of its two-sided 99\,\% Clopper--Pearson interval. For C2 and C4 the archived training logs list runs more than once (a second re-evaluation pass was appended); for C4 this affects a third of the rows. $n$ counts de-duplicated runs (C2: 286 rows $\to$ 282 runs, C4: 108 rows $\to$ 71 runs).}
\label{tab:oneshot-risk}
\centering\small
\begin{tabular}{lllrllllr}
\toprule
Config. & Set & Split & $n$ & Median & $q_{0.05}$ & $q_{0.05}\,/\,\mathrm{chance}$ & $p$ ($t$-test) & $\hat{p}_{>\mathrm{chance}}$ (CI$_{99}$) \\
\midrule
\multicolumn{9}{l}{\emph{Historic mine, training campaign, structural splits ($K=95$)}} \\
C1 & $F_V^{\mathrm{a}}$ & device & 299 & 0.953 & 0.949 & 90.1 & $<10^{-16}$ & 299/299 ($\geq$0.982) \\
C2 & $F_{HZ}^{\mathrm{a}}$ & subject$^{\dagger}$ & 282 & 0.943 & 0.932 & 88.6 & $<10^{-16}$ & 282/282 ($\geq$0.981) \\
C3 & $F_{HZ}^{\mathrm{b}}$ & device & 299 & 0.882 & 0.874 & 83.1 & $<10^{-16}$ & 299/299 ($\geq$0.982) \\
C4 & $F_V^{\mathrm{b}}$ & subject & 71 & 0.666 & 0.654 & 62.1 & $<10^{-16}$ & 71/71 ($\geq$0.928) \\
C5 & $F_K$ & device & 77 & 0.422 & $3\times10^{-4}$ & 0.03 & $1\times10^{-13}$ & 43/77 ($\geq$0.406) \\
\addlinespace
\multicolumn{9}{l}{\emph{Same models, re-evaluated at the re-measurement, present classes ($K=42$)}} \\
C1 & $F_V^{\mathrm{a}}$ & device & 299 & 0.407 & 0.392 & 16.5 & $<10^{-16}$ & 299/299 ($\geq$0.982) \\
C2 & $F_{HZ}^{\mathrm{a}}$ & subject$^{\dagger}$ & 282 & 0.370 & 0.355 & 14.9 & $<10^{-16}$ & 282/282 ($\geq$0.981) \\
C3 & $F_{HZ}^{\mathrm{b}}$ & device & 299 & 0.353 & 0.338 & 14.2 & $<10^{-16}$ & 299/299 ($\geq$0.982) \\
C4 & $F_V^{\mathrm{b}}$ & subject & 71 & 0.430 & 0.416 & 17.5 & $<10^{-16}$ & 71/71 ($\geq$0.928) \\
C5 & $F_K$ & device & 77 & 0.318 & $9\times10^{-4}$ & 0.04 & $9\times10^{-13}$ & 43/77 ($\geq$0.406) \\
\addlinespace
\multicolumn{9}{l}{\emph{Uranium mine ($K=40$)}} \\
U1 & $F_V^{\mathrm{a}}$ & device & 78 & 0.648 & 0.640 & 25.6 & $<10^{-16}$ & 78/78 ($\geq$0.934) \\
U2 & $F_V^{\mathrm{b}}$ & device & 24 & 0.649 & 0.639 & 25.6 & $<10^{-16}$ & 24/24 ($\geq$0.802) \\
U3 & $F_{HZ}$ & device & 9 & 0.625 & 0.616 & 24.6 & $<10^{-16}$ & 9/9 ($\geq$0.555) \\
\bottomrule
\end{tabular}

\par\smallskip\noindent\footnotesize $^{\dagger}$~The validation assignment of this repeat batch cannot be reconstructed from the logs. It is not to be read as a statement about cross-subject performance (section~\ref{sec:oneshot-risk}).
\end{table}

\subsection{Retrospective verdicts}\label{sec:verdictsub}

Table~\ref{tab:verdicts} applies the criteria retrospectively to all stages and both mines with $k=5$ and $k=10$. All deployment stages pass both levels. The margins order the stages: in frame A the cross-device ensemble exceeds the $k{=}10$ threshold by a factor of 9.0 ($q_{0.05}=0.949$ against $10/95=0.105$), the cross-subject ensemble by 6.2 ($n=71$ de-duplicated repeat runs), and the class-exact cross-time ensemble by 1.6 ($q_{0.05}=0.392$ against $10/42=0.238$). The criteria remain binding at the stage with the hardest shift.

Two observations qualify the verdicts. First, the neighbour-tolerant metric needs its own chance reference, because a tolerated set spans more than one class. In the implementation used, the tolerated set of a class consists of the class itself and of every direction-resolved class of the segments adjacent on the gallery adjacency graph; the opposite direction of the class's own segment does not count as a hit. Over the 42 present classes it averages 6.81 classes, so a uniform random predictor scores $0.162$. Against this reference the neighbour-tolerant $q_{0.05}=0.612$ corresponds to a factor of $3.8$, and a $k{=}5$ criterion is not met. The $F_A$ family is no empirical counter-check here: it predicts a single class only, so its value of $0.155$ (table~\ref{tab:stage4-longtime}) is the value of that one class under the averaging convention of the archived pipeline. Under the convention of equation~(\ref{eq:pmac}), which counts never-predicted classes as zero, the same value drops to $0.004$. The neighbour-tolerant metric therefore enters the protocol as descriptive evidence with its own declared reference, and the binding roll-out criterion is the class-exact metric. Second, in frame B the repeat ensemble of the cross-device candidate has a median of 0.648 with a tight spread ($q_{0.05}=0.640$, $q_{0.95}=0.654$), while the single screening run of the identical configuration reports 0.825, far outside the ensemble support. The gap of 0.177 is twelve times the entire ensemble width, and the second candidate configuration of the same frame shows the same gap (0.826 against $q_{0.05}=0.639$). For the structurally identical case C2 in frame A, a joint re-scoring of all models on identical validation data established training stochasticity as the cause; for frame B that control is missing, because the validation assignment there cannot be reconstructed. Whether the gap is therefore due to training stochasticity or to an unlogged difference in the evaluation set-up cannot be decided for frame B. For the conclusion drawn here that is immaterial: in either case the single screening value is not defensible as acceptance evidence.

\noindent$\Rightarrow$\emph{A single training run is therefore unsuitable as acceptance evidence, and this is why the decision rule operates on repeat ensembles and quantiles.}

\begin{table}
\caption{\label{tab:verdicts}Retrospective application of the acceptance criteria to all stages and both mines. Basis is the repeat ensemble of the candidate configuration ($n$ runs); for the cross-subject stage in mine B no repeat ensemble exists and the 5\,\% quantile over the $F_V$ family screening runs is used as a conservative surrogate. Thresholds: $5/K$ and $10/K$ with $K=95$ (A), $K=40$ (B), $K_{\mathrm{p}}=42$ (cross-time, present-conditioned).}
\centering\footnotesize\setlength{\tabcolsep}{3.5pt}
\begin{tabular}{@{}llllllll}
\toprule
Mine & Stage & Metric & $n$ & median & $q_{0.05}$ & $q_{0.95}(\mathrm{OOPS})$ & verdict $k{=}5$ / $k{=}10$ \\
\midrule
A & Random split & class-exact & 112 & 0.992 & -- & -- & upper bound only \\
A & Cross-device & class-exact & 299 & 0.953 & 0.949 & -- & pass / pass \\
A & Cross-subject & class-exact & 71 & 0.666 & 0.654 & -- & pass / pass \\
A & Cross-time & class-exact (present) & 299 & 0.407 & 0.392 & 0.198 & pass / pass \\
A & Cross-time & neighbour-tolerant$^{\rm b}$ & 299 & 0.632 & 0.612 & 0.198 & descriptive$^{\rm b}$ \\
B & Random split & class-exact & 1185 & 0.982 & -- & -- & upper bound only \\
B & Cross-device & class-exact & 78 & 0.648 & 0.640 & -- & pass / pass \\
B & Cross-subject & class-exact & 144\,$^{\rm a}$ & 0.465 & 0.392 & -- & pass / pass \\
\bottomrule
\end{tabular}

\par\smallskip\noindent\footnotesize $^{\rm a}$ surrogate: family screening runs including hyperparameter spread. Random-split rows report the median over screening runs (A: the 112 matched configurations, B: all 1185 screening runs); they enter the protocol as the declared upper bound and receive no roll-out verdict. Cross-time rows use the device-split candidate ensemble; the subject-split ensemble gives $q_{0.05}=0.416$ (class-exact), $0.615$ (neighbour-tolerant), $q_{0.95}(\mathrm{OOPS})=0.191$, with identical verdicts. $^{\rm b}$ the neighbour-tolerant row is descriptive: its chance reference is the mean prevalence of the tolerated sets ($6.81/42=0.162$ on the adjacency graph), and no $k$-verdict is assigned (section~\ref{sec:verdictsub}).

\end{table}

\subsection{Cost of adoption}
\label{sec:cost}

Table~\ref{tab:cost} summarises the training compute per stage from the logged per-run training times. The full two-mine study comprises 7003 training runs and 3237 GPU\,h. A minimal adoption of the protocol needs a small fraction of this. It consists of one candidate configuration, $n$ repeat training runs under the cross-device split (median 1342\,s per run, interquartile range 1168 to 1466\,s) and one cross-time repetition, which re-evaluates the stored models on later recordings and adds no training compute. A guarded quantile verdict requires $n=59$ and costs 22.0 GPU\,h, that is 0.68\,\% of the full study. A mere point estimate of the quantile needs $n=30$ at 11.2 GPU\,h (0.35\,\%), a screening $n=10$ at 3.7 GPU\,h.

\begin{table}
\caption{\label{tab:cost}Training compute per protocol stage (from logged per-run training times on the study GPUs) and the minimal adoption variant. The cross-time stage adds no training compute because it re-evaluates existing models.}
\centering\small
\begin{tabular}{@{}llrrr}
\toprule
Frame & Stage & runs & median s/run & GPU\,h \\
\midrule
A & random screening & 1306 & 247 & 202 \\
A & cross-device & 416 & 546 & 114 \\
A & cross-subject & 504 & 351 & 138 \\
A & repeat ensembles & 1072 & 1412 & 908 \\
B & random screening & 1185 & 1059 & 578 \\
B & cross-device & 1152 & 558 & 644 \\
B & cross-subject & 1152 & 459 & 564 \\
B & repeat ensembles & 216 & 941 & 90 \\
\midrule
\multicolumn{2}{@{}l}{full study} & 7003 & & 3237 \\
\multicolumn{2}{@{}l}{minimal protocol, $n{=}10$} & 10 & 1342 & 3.7 \\
\multicolumn{2}{@{}l}{minimal protocol, $n{=}30$} & 30 & 1342 & 11.2 \\
\multicolumn{2}{@{}l}{minimal protocol, $n{=}59$ (guarded)} & 59 & 1342 & 22.0 \\
\bottomrule
\end{tabular}

\par\smallskip\noindent\footnotesize Minimal protocol: one candidate configuration ($F_V$, $L=700$, step size 1, hidden size 512, 2 layers, batch 256, learning rate $10^{-3}$), $n$ repeat training runs under the cross-device split, plus one cross-time repetition by re-evaluation. At $n=30$ this is 0.35\,\% of the full study's training compute, at $n=59$ it is 0.68\,\%. Only from $n=59$ does a valid non-parametric 95\,\% lower bound for the 5\,\% quantile exist (section~\ref{sec:oneshot-risk}). Row values are rounded; the exact study total is 3237.5 GPU\,h.

\end{table}

\section{Discussion}
\label{sec:discussion}

\subsection{What the protocol measures and what it changes}

The staged design makes two effects measurable that an evaluation within a single campaign does not resolve. The first is the optimism of the customary random split. Paired per configuration, it amounts to $+0.14$ under device change and $+0.41$ under subject change, both with a 95\,\% bootstrap confidence interval. The second is the change of the magnetic signatures over time. Of the apparent 47-point loss, the largest part is accounted for by the class inventory and by annotation boundaries, and for the genuine signature change a defensible upper bound of 23 points remains. The most consequential single observation for practice concerns the ranking: the representation ranking of the random split does not predict the deployment ranking ($\tau\le0$ against the cross-subject stage in both mines, indistinguishable from zero in every comparison), while the two deployment stages lie closer together. With eight feature sets this rules out strong rank agreement, not moderate agreement (section~\ref{sec:representations}). Model selection on random-split scores therefore selects for the wrong regime, and the selection error is invisible without the staged design. The foundational paper \cite{platte2026companion} selects hyperparameters inside the same model clouds by attribution; the protocol here decides which stage's scores that selection may read.

The cross-time stage adds an observation with operational value: models trained under the harder structural splits in the training campaign age no worse. A genuine advantage, by contrast, cannot be evidenced. Over the 112 matched triples the margin against random-split training is $+0.004$, its confidence interval includes zero, and it lies below the $+0.007$ produced by the residual learning-rate-schedule difference alone (section~\ref{sec:optimism_gap}). The cell-median gap of ten points does not measure ageing in any case but the differing composition of the cells (section~\ref{sec:stage4-longtime}). What the protocol achieves here is the separation of these three quantities; single-campaign evaluation cannot keep them apart.

\subsection{Transferability beyond the case study}

The protocol was \emph{abstracted from magnetometry.} Its inputs are a class inventory, a device fleet, personnel, and recording epochs; its instruments are matched-configuration pairing, chance references with correct class counts, present-conditioning with an OOPS guard, neighbour tolerance where classes have a spatial or ordinal adjacency, and repeat ensembles with quantile verdicts. Device fleets that age, personnel that changes and class inventories that drift are standard conditions of industrial sensor systems, from inertial wearables in occupational safety to condition monitoring across machine generations. The minimal adoption in table~\ref{tab:cost} was costed for exactly this transfer situation: one candidate configuration, 59 repeat trainings under a device-held-out split and one re-evaluation on later recordings, at $0.68\%$ of the compute of the full study reported here.

The protocol also connects to a regulatory development. For high-risk AI systems the EU AI Act \cite{EuropeanParliamentandCounciloftheEuropeanUnion_2024_RegulationEU20241689} requires a risk management system with testing against defined metrics and probabilistic thresholds (Article~9), declared levels of accuracy and robustness in the instructions for use (Article~15), and post-market monitoring over the operating period (Article~72). The staged verdicts, the declared reference levels and the cross-time stage supply measurable counterparts for these obligations. We state this as an anchoring, and a legal assessment is outside the scope of this paper.

\subsection{Uncertainty budget}
\label{sec:budget}

Table~\ref{tab:budget} consolidates the uncertainty sources of the protocol's results, separating what this study quantifies from what it can only bound or declare. The quantified components follow from the repeat ensembles and the control computations of sections~\ref{sec:stage4-longtime} and~\ref{sec:uncertainty}. Three sources remain unquantified and are declared as such. The cross-time validation set consists of two walks of the accessible loop, so sampling uncertainty at walk level rests on $n=2$ and is not estimable from within the data; the sensor calibration state of the individual magnetometers between campaigns was not tracked; and the label timing of the training campaigns carries the 2\,s propagation tolerance of the annotation transfer to the secondary devices together with the changed marker-switch convention between the campaigns. Secular and diurnal variation of the geomagnetic field over the 34 months is a candidate contributor to the residual of the decomposition and is not separable from it here.

\begin{table}
\caption{\label{tab:budget}Uncertainty budget of the protocol results. Quantified components are stated with their observed magnitude on the macro-precision scale; declared components are named with the reason they remain unquantified.}
\centering\small
\begin{tabular}{@{}p{0.30\linewidth}p{0.17\linewidth}p{0.44\linewidth}@{}}
\toprule
Source & Status & Magnitude / basis \\
\midrule
Training stochasticity & quantified & repeat-ensemble quantiles; IQR $0.012$--$0.015$
for stable configurations, bimodal for one $F_K$ configuration
(section~\ref{sec:uncertainty}) \\
Evaluation-pipeline reconstruction & quantified & $\le 1.2$\,pp, control re-scoring of
79 archived models (section~\ref{sec:stage4-longtime}) \\
Class-inventory change & exact & multiplicative identity $42/95$, reproduced to
$3\times10^{-16}$ \\
Annotation-boundary ambiguity & bounded & $16.5$--$25.8$\,pp recovered by one-neighbour
tolerance (seven learned feature sets; negative control $F_A$ excluded). About half of that
is reached by a chance predictor as well (section~\ref{sec:stage4-longtime}) \\
Optimism of the random split & quantified & $+0.14$ / $+0.41$ paired median bias
(section~\ref{sec:optimism_gap}) \\
Scheduler confound of the gap & bounded & $\le0.007$ median (35 matched pairs,
section~\ref{sec:optimism_gap}) \\
Composition of the measurand & quantified & bootstrap over the 42 reachable location
classes: 95\,\% interval $[0.33, 0.48]$ for the cross-time candidate C1 and $[0.35, 0.51]$
for C4 (class spread $\mathrm{SD}=0.25$, three classes at precision zero). This is the
dominant component, an order of magnitude wider than training stochasticity \\
$q_{0.05}$ estimator (order statistics) & bounded & rests on the 4th--16th order statistic
for $n\ge71$; a valid non-parametric 95\,\% lower bound for $q_{0.05}$ exists only from
$n=59$ (section~\ref{sec:oneshot-risk}) \\
Test-set sampling (cross-time) & declared & two walks of the loop; walk-level $n=2$,
not estimable from within the data at that level \\
Sensor calibration state & declared & per-device calibration between campaigns not
tracked; datasheet error bounds $\pm(\qtyrange[range-phrase = --,range-units = single]{0.1}{0.45}{\micro\tesla})$ \\
Label timing (training campaigns) & declared & $2$\,s propagation tolerance;
marker-switch convention differs between campaigns \\
Geomagnetic field variation & declared & secular and diurnal variation contained in,
and not separable from, the decomposition residual \\
\bottomrule
\end{tabular}
\end{table}

\subsection{Limitations}

The evidence base has known boundaries. The cross-subject axis rests on two persons per mine, so subject effects are demonstrated but not sampled as a population; the protocol certifies robustness against training stochasticity and declares no population inference. The 112 matched triples behind the optimism gap share all varied hyperparameters except the learning-rate schedule, which differed between the screening and the structural campaigns; the gap is therefore a field observation on matched configurations rather than a controlled ablation. The markers of the re-measurement were reconstructed from planning documents after the physical markers had been removed, which is precisely why the neighbour-tolerant metric exists; the decomposition quantifies this artefact rather than assuming it away. The two sites cover two magnetic distortion regimes, narrow ore-vein galleries and large gallery cross-sections, and further regimes remain to be tested. In the uranium mine, device and person are moreover firmly coupled: each operator carried the same devices throughout the campaign, so a device change across the carrier boundary is at the same time a change of person. Which device the splits there held out is not logged. The cross-device figures of that site are therefore to be read as an upper bound on a pure device effect. Finally, the repeat ensembles share one data split per campaign; they quantify training stochasticity, and campaign-level resampling of routes or days is future work.

\section{Conclusion}
\label{sec:conclusion}

We have treated the evaluation of a deployed sensor-AI system as a measurement and supplied the elements such a measurement owes its reader: a staged design that holds out devices, subjects and time; declared reference levels including a demoted random split; an uncertainty statement built from repeat trainings; and an explicit decision rule whose verdicts were applied retrospectively to every stage of two mines, including one sensitivity case in which the criterion is not met. The demonstration yields concrete, transferable practice:

\begin{enumerate}
\item Report the random split only as a declared upper bound, together with its window overlap. In matched configurations it overstated deployment performance by up to $+0.41$ and its ranking did not predict the deployment ranking.
\item Judge roll-out on quantiles of repeat ensembles against chance references with the correct class count. A mean-based test accepted a configuration whose $5\%$ quantile lies more than an order of magnitude below chance.
\item Condition long-term scores on the present class inventory and guard them with the OOPS rate. One representation family combined competitive present-conditioned scores with three to four times the misdirection rate of its alternatives.
\item Decompose long-term degradation before attributing it to the phenomenon. Of an apparent 47-point loss, class loss alone explained 23 points, and a defensible upper bound of 23 points remained for genuine signature change.
\item Select models under the shift that deployment will actually impose rather than relying on a ranking from the random split. Our data rule out a strong agreement between the two rankings; over 34 months, models from the harder regimes at any rate aged no faster.
\end{enumerate}

The full study cost 3237 GPU-hours; adopting the protocol for a new system costs a fraction of a percent of that. The instrument the paper offers is the staged report itself: a deploying actor who presents these verdicts to a reviewing forum presents evidence rather than a single flattering number.

\section*{Author contributions}
Roles follow the CRediT taxonomy (ANSI/NISO Z39.104-2022). \textbf{Benny Platte:} conceptualization, methodology, software, formal analysis, investigation, data curation, visualization, writing --- original draft. \textbf{Rico Thomanek:} conceptualization, investigation. \textbf{Christian Roschke:} project administration. \textbf{Marc Ritter:} supervision.

\ack{The authors thank the operators of the Markus-R\"ohling-Stolln visitor mine and Wismut GmbH for access to the underground sites.}

\data{The training-log data that support the findings of this study, together with the analysis scripts that reproduce every figure and table, are available from the corresponding author upon reasonable request. For the re-measurement, per-run confusion matrices are available and every metric reported from it was recomputed from them. For the training campaign in the historic mine no validation confusion matrices are archived; the metrics there come from the logged metric columns and were checked, wherever the same models recur in the re-measurement, against that recomputation. Two derived summary files of the source archive carried undocumented scale factors on the aggregate metrics of one feature set; these were identified during the recomputation and reversed, and the corrected values were verified against the confusion matrices to numerical precision (section~\ref{sec:protocol:training}).}

\bibliographystyle{unsrtnat}
\bibliography{literatur,eigene}

\end{document}